\documentclass[10pt]{article}
\usepackage{float}
\usepackage[margin=1in]{geometry}
\usepackage{amsmath,amsthm,amssymb}
\usepackage{graphicx}
\usepackage{titlesec}
\usepackage{natbib}
\usepackage{authblk}
\usepackage{caption}
\usepackage{subfig}
\usepackage{textcomp}
\usepackage{enumerate}
\usepackage{tabularx,ragged2e}
\usepackage{multirow}
\usepackage{color}
\usepackage{diagbox}
\usepackage[utf8]{inputenc}
\usepackage{threeparttable}
\usepackage{tabularx,ragged2e}
\usepackage{txfonts}
\usepackage{caption}
\usepackage{booktabs}
\usepackage{stfloats}
\usepackage{makecell}
\usepackage{multicol}
\definecolor{red}{rgb}{1,0,0}
\definecolor{blue}{rgb}{0,0,1}
\usepackage{algorithm}
\usepackage{algorithmicx}
\usepackage{algpseudocode}  
\usepackage{lineno} \linenumbers
\usepackage{amsfonts}
\usepackage{flushend}
\usepackage{balance}
\usepackage{longtable}   
\usepackage{bm}
\usepackage{url}

\newcommand{\del}[1]{}

\renewcommand\baselinestretch{2}

\nolinenumbers

\titleformat*{\section}{\large\bfseries}
\titleformat*{\subsection}{\normalsize\bfseries}
\titleformat*{\subsubsection}{\small\bfseries}

\begin{document}

\title{CryoAlign: feature-based method for global and local 3D alignment of EM density maps}
\renewcommand\footnotemark{}
\author{Bintao He$^{1,\dagger}$, Fa Zhang$^{2,\dagger}$, Chenjie Feng$^{3}$, Jianyi Yang$^{1}$, Xin Gao$^{4,*}$, Renmin Han$^{1,*}$ \thanks{$^{\dagger}$These authors should be regarded as Joint First Authors; $^*$All correspondence should be addressed to Xin Gao (xin.gao@kaust.edu.sa) and Renmin Han (hanrenmin@sdu.edu.cn).}}
\date{\small $^1$Research Center for Mathematics and Interdisciplinary Sciences, Shandong University, Qingdao 266237, China; $^2$School of Medical Technology, Beijing Institute of Technology; $^3$College of Medical Information and Engineering, Ningxia Medical University, Yinchuan 750004, China; $^4$King Abdullah University of Science and Technology (KAUST), Computational Bioscience Research Center (CBRC), Computer, Electrical and Mathematical Sciences and Engineering (CEMSE) Division, Thuwal, 23955, Saudi Arabia.}

\maketitle

\renewcommand\baselinestretch{2}

\vspace{-2em}
\begin{abstract}

Advances on cryo-electron imaging technologies have led to a rapidly increasing number of density maps. Alignment and comparison of density maps play a crucial role in interpreting structural information, such as conformational heterogeneity analysis using global alignment and atomic model assembly through local alignment. Here, we propose a fast and accurate global and local cryo-electron microscopy density map alignment method CryoAlign, which leverages local density feature descriptors to capture spatial structure similarities. CryoAlign is the first feature-based EM map alignment tool, in which the employment of feature-based architecture enables the rapid establishment of point pair correspondences and robust estimation of alignment parameters. Extensive experimental evaluations demonstrate the superiority of CryoAlign over the existing methods in both alignment accuracy and speed.
\end{abstract}

\section{Background}
Density maps obtained through cryo-electron microscopy (EM) provide key information for protein structure determination and function analysis \cite[]{bai2015cryo, nogales2016development}. The Electron Microscopy Data Bank \cite[]{lawson2016emdatabank}, a public database, has accumulated more than twenty thousand entries by October 2022. With technological advancement, new entries with fine resolution (10\text{\AA} or better) dominate recent years, accounting for more than 90 percent of the database. To extract and interpret the underlying structural information from EM density maps, there is a strong demand for fast and accurate alignment and comparison of EM maps, especially for entries with high resolution. For example, comparison of superimposed density maps helps to identify variable areas associated with heterogeneity or explore structural changes of different function states \cite[]{poitevin2020structural, rheinberger2018ligand, joseph2020comparing}. In protein macromolecular complex modelling, accurate local alignment effectively accelerates the chain assembly process \cite[]{he2022model, woetzel2011bcl, zhang2022cr, van2015fast}. Additionally, similarity scores derived from alignment can serve as feasible metrics for EM map retrieval problems \cite[]{joseph2017improved, farabella2015tempy}. However,  density maps with high and medium resolutions contain a great amount of rich and clear structural information, raising high requirements on alignment accuracy and efficiency.

Several works have been developed to address the EM map alignment problem. Gmfit \cite[]{suzuki2016omokage,kawabata2008multiple} represents the EM density maps with Gaussian mixture models (GMM) and utilizes maximization of correlation between the Gaussian functions to optimize the global transformation parameters. The balance between speed and approximation accuracy of GMM is determined by the used number of Gaussian kernels. Gmfit utilizes a combination of Gaussian functions far less than the total number of raw atoms to represent a map, providing fast and robust, but less accurate alignment results, which makes gmfit more suitable for low-resolution maps. Chimera, a widely used software for molecular manipulation and visualization, offers a map fitting method known as fitmap \cite[]{pettersen2004ucsf}. Fitmap directly performs local optimization to maximize the correlation between maps, starting from multiple randomly generated initial placements of the source map. However, due to the significant influence of the initial location of the maps, fitmap usually needs users' intervention or the use of preset locations to achieve satisfactory results. Recently, a vector-based EM density map alignment method called VESPER \cite[]{han2021vesper} was proposed for both better alignment and retrieval performance. VESPER utilizes a collection of vectors that are specifically oriented toward the local density maximum 
to capture the intricate 3D structures embedded in the maps \cite[]{terashi2018novo}. Using the sum of dot products between matched vectors from two maps, VESPER finds the best alignment parameters by exhaustive search of both rotational and translational intervals. Compared to gmfit and fitmap, the point distribution retains abundant information about spatial structures and the orientations of vectors explicitly depict local density trends.
However, the parameter optimization of VESPER is based on an exhaustive search on spacial rotation and translation with a given search interval, which leads to inflexible and insufficient optimization and large execution time.

Here, we proposed a global and local EM density map alignment method, CryoAlign, to achieve fast, accurate and robust comparison of two EM density maps by utilizing local spatial feature descriptors. In CryoAlign, the density map is sampled to generate a point cloud representation, and a clustering process is applied to the point cloud to extract key points based on local properties such as the density value distribution and the connectivity of points. Once the key points are identified, CryoAlign calculates local feature descriptors by collecting the distribution of density directions in the vicinity. These feature descriptors capture rich information about the local structural characteristics of the density map, significantly reducing the number of points to be considered, and leading to a more efficient alignment process. Meanwhile, the local feature descriptors computed based on the distribution of density directions provide a comprehensive representation of the local structural variations. Using these feature descriptors, CryoAlign employs a mutual feature matching strategy to establish correspondences between keypoints in different density maps, allowing for stable alignment parameter estimation. To further refine the alignment, CryoAlign applies a point-based iterative method, aiming to bring overlapping point pairs closer together. 

To assess the performance of CryoAlign, comprehensive evaluations are conducted on diverse test sets, which demonstrate its high alignment accuracy for both global and local EM map alignment. In comparison to other alignment methods such as gmfit, fitmap, and VESPER, CryoAlign stands out by providing more precise superimposition of density maps while maintaining a lower failure ratio. 

\section{Results}

\begin{figure}
	\centering
	\includegraphics[width=0.9 \textwidth]{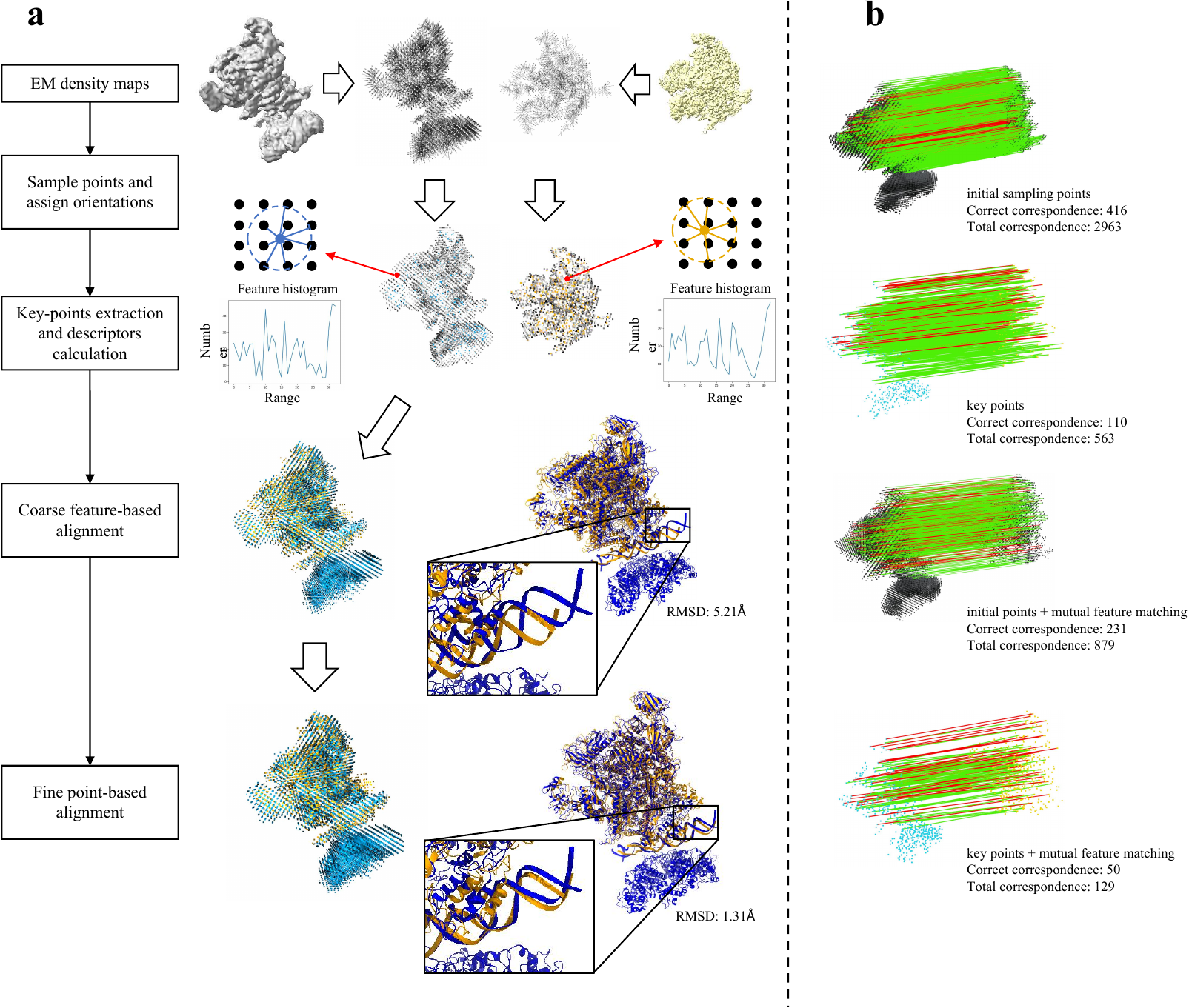}
	\caption{\textbf{Overview of CryoAlign. a} The flowchart of CryoAlign. A visual example of RNA polymerase-sigma54 holo enzyme and promoter DNA closed complex with (EMD-3696, left) and without (EMD-3695, right) transcription activator PspF intermedate is provided on the right. The inputs of CryoAlign are a pair of EM density maps. First, initial point clouds are sampled by given interval and density vectors are computed for all points. Then, clustering algorithms are applied to extract key points that represent the rough backbones of the structures. Local spatial structural feature descriptors are calculated to capture the local structures around these key points. Using the extracted feature descriptors and the mutual feature matching technique, CryoAlign robustly and efficiently computes the initial pose parameters. Finally, CryoAlign generates the best superimposition by iteratively making the corresponding points closer together. The alignment parameters are then applied to the fitted atom models, directly illustrating the alignment performance. \textbf{b} The proportion of correct correspondences. In the visual example, the lines between points represent the estimated correspondences, with correct correspondences labeled in red and false ones labeled in green. From top to bottom, four cases with only initial points, with only extracted key points, with combination of initial points and mutual matching, and with combination of key points and mutual matching are listed.}
	\label{fig:overview}
\end{figure}
\subsection{Overview of the CryoAlign procedure}
Figure \ref{fig:overview}a illustrates the workflow of CryoAlign. When provided with a density map, CryoAlign applies a uniform sampling to generate a set of initial grid points, which act as the starting positions for the subsequent alignment process. At each sampled grid point, a corresponding density vector is assigned to reflect the trend of changes in density within its vicinity. These density vectors are derived from VESPER, which demonstrate their effectiveness as a representation of the local density variations around the grid points. However, the excessive number of initial grid points and the limited representation range of density vectors make them unsuitable for direct alignment. Then, CryoAlign uses mean shift algorithm\cite[]{carreira2006acceleration} to identify local dense points and apply density-based spatial clustering method\cite[]{ester1996density} to find cluster centers as the key points of point clouds (see the "Methods" section). The key points extracted in CryoAlign are chosen to consider both the distribution of density values and the connectivity of points, providing a rough representation of the protein backbones. Local spatial structural feature descriptors are calculated on the extracted key points in CryoAlign by analyzing the directions and distribution of density vectors within their vicinity. Compared to vectors, local feature descriptors capture structural information from multiple neighboring points instead of just a single grid point. This approach provides a more distinctive and comprehensive description of the local region, effectively improving the accuracy of the alignment results. Next, CryoAlign implements a two-stage alignment approach to achieve an accurate superimposition from coarse to fine. In the first stage, CryoAlign utilizes a mutual feature matching strategy to establish correspondences between key points and efficiently estimate initial poses. This stage allows fast and stable alignment, laying the foundation for subsequent refinement. In the second stage, CryoAlign focuses on achieving the best possible superimposition. It considers the point-to-point correspondences between the initial grid points and employs an iterative process to bring these points closer together. By iteratively adjusting the positions of overlapping points, CryoAlign continues to improve the alignment and strives for optimal alignment accuracy.

For a more illustrative explanation, a visual example of local alignment is provided on the right side of Figure \ref{fig:overview}. The two input EM maps in the example represent the structure of RNA polymerase-sigma54 holoenzyme with promoter DNA closed complex. However, there is an additional transcription activator PspF intermediate present in the left map (EMD-3696), while it is absent in the right one (EMD-3695). The top row of the visual example displays the grid-sampled point clouds of the two maps, represented by dark points, along with their corresponding vectors. These vectors effectively preserve the three-dimensional volumes and shapes of the structures. The second row showcases the extracted key points, represented by colored points, and presents an example of a spatial structural feature histogram pair. These feature histogram pairs are used for alignment by filtering and selecting the most relevant and informative feature pairs. Following the direction of the hollow arrows, the two point clouds are aligned based on the filtered feature pairs. The coarse alignment stage provides an initial alignment that is approximately correct, although imperfect, with a high degree of overlap between the structures. Subsequently, the point-based stage is employed to refine the alignment and achieve the best possible superimposition by minimizing the distances between corresponding point pairs. Furthermore, for better visual evaluation, the corresponding PDB atom structures transformed by the alignment parameters are also attached in the example.

\subsection{Datasets of density maps and metrics}
\textbf{Datasets.} To evaluate the performance of global and local alignment, we utilize the EM maps from the datasets provided by VESPER, which are specifically designed for global and local density map search. We begin by filtering maps without fitted PDB atom models\cite[]{berman2000protein} and focus on collecting the maps with a resolution higher than 10\text{\AA}. As a result of the filtering process, we obtain two datasets for evaluation: the global alignment dataset, which consists of 64 pairs of EM maps, and the local alignment dataset, which contains 201 map pairs. In Table \ref{tab:dataset_statistic}, we present the statistical information for these map pairs. The first column, labeled "Res. range," indicates the resolution range of the inputted maps. The column labeled "Cross res." indicates whether the inputted pairs are from different resolution ranges. Using these two datasets, we assess the performance of our alignment method indirectly by analyzing the fitted PDB models, both quantitatively and qualitatively. 
Furthermore, to evaluate the algorithm's performance in atomic model fitting, we also utilize intermediate-resolution protein complexes datasets provided by \cite[]{he2022model}. We select eight protein complexes of 4.0$\sim$8.0\AA, and each has 2$\sim$5 single chains. By leveraging these diverse datasets, we are able to comprehensively evaluate the alignment performance of our method across different scenarios and applications.

\begin{table}[h]
	\caption{Resolution statistical distribution of datasets}
	\label{tab:dataset_statistic}
	\centering
	\setlength{\tabcolsep}{4.mm}{
		\begin{threeparttable}
		\begin{tabular}{l|cc}
			\toprule
			Res. range   & global alignment & local alignment \\
			\midrule
			\textless{}5\AA & 35               & 122             \\
			5$\sim$10\AA    & 16               & 14              \\
			Cross res.        & 13               & 65              \\
			\bottomrule
		\end{tabular}
		\begin{tablenotes}
			\footnotesize
			\item Res. range, resolution range. The "Cross res." means that the inputted pairs are from different resolution ranges. The number of density maps is counted based on the resolution range.
		\end{tablenotes}
		\end{threeparttable}
	}
\end{table}

\textbf{Alignment metric.} To quantitatively evaluate the alignment performances, ground truth for the superimposition is defined by computing the transformation parameters using MM-align\cite[]{mukherjee2009mm} on fitted atom models. We then calculate the root mean square distances (RMSD) between the ground truth and the alignment results obtained by different methods. It is important to note that we consider an alignment as a failure if the RMSD exceeds 10Å. This threshold helps identify cases where the alignment deviates significantly from the ground truth. Additionally, to provide a more intuitive visualization, the fitted PDB structures are transformed using the alignment parameters, allowing a direct comparison of the aligned structures.

\subsection{Global alignment accuracy}
\begin{figure}[H]
	\centering
	\includegraphics[width=0.85 \textwidth]{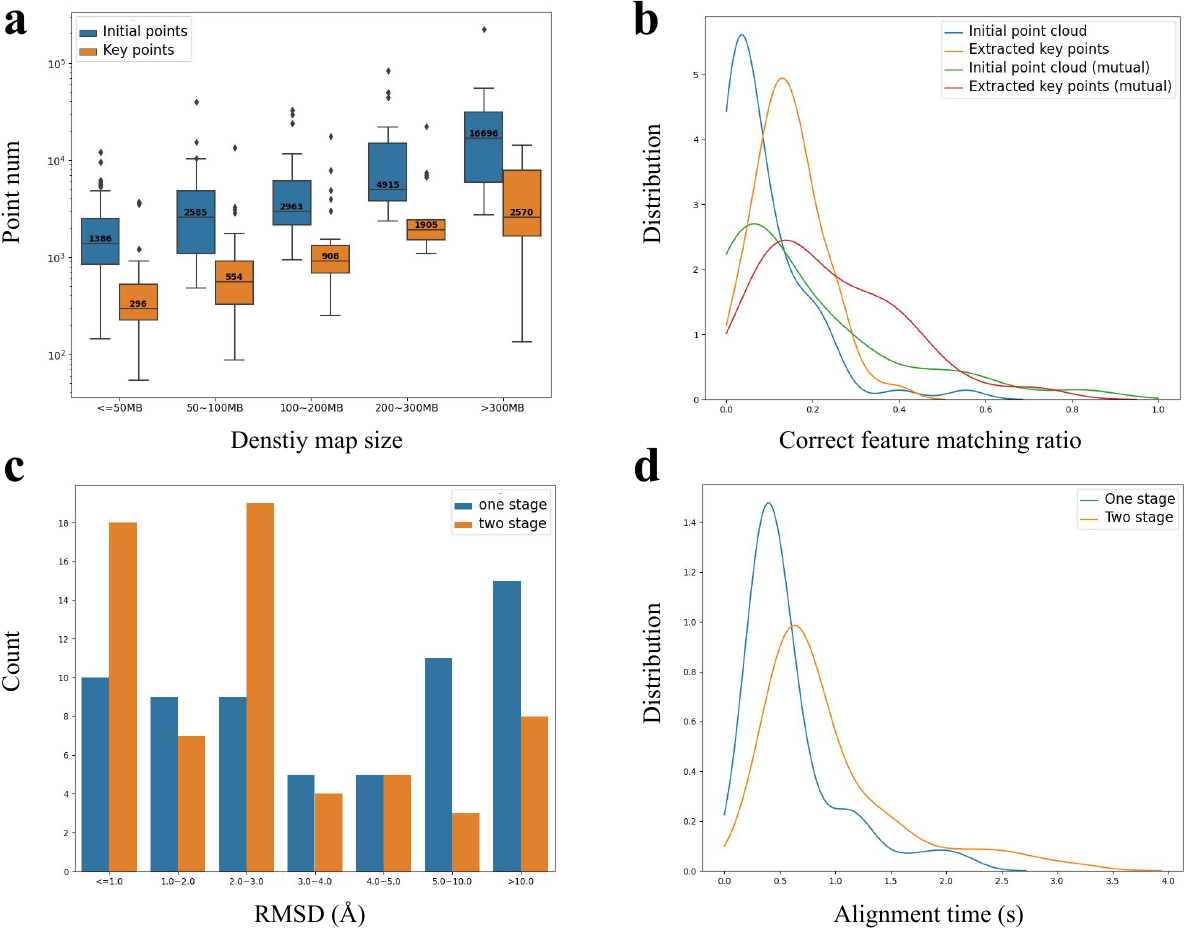}
	\caption{\textbf{Global alignment performance of CryoAlign. a} The number of points with the increasing density map size. Initial points (blue box) and key points (orange box). \textbf{b} The correct ratio distribution of four different feature matching strategies. Only initial points (blue line), only key points (orange line), initial points + mutual matching (green line) and key points + mutual matching (red line). \textbf{c} The accuracy distribution of one stage alignment and two-stage alignment. \textbf{d} The execution time of one stage alignment and two-stage alignment.}
	\label{fig:global_performance}
\end{figure}

First, we thoroughly assess the alignment performance of CryoAlign in the pre-collected dataset. We initially sample the density maps with an interval of 5\text{\AA}, which provide sufficient spatial distribution information for global alignment. Figure \ref{fig:global_performance}a shows the mean number of initial sampling points and extracted key points as the size of inputted density maps increases. After key points extraction, the point clouds typically reduce in size to around $10\%\sim20\%$ of initial points, making subsequent calculations more efficient. Furthermore, the key points effectively represent the protein backbones, leading to more accurate feature correspondence establishment. Figure \ref{fig:global_performance}b presents a comparison of the feature correspondence accuracy under different scenarios, including different point cloud representations and the utilization of mutual feature matching. The orange and red curves are consistently positioned to the right of the other two curves, indicating that the utilization of key points can mitigate feature mismatches caused by excessive sampling. Additionally, the mutual feature matching strategy considers point pairs that are closest to each other in the feature domain, further enhancing the accuracy of correspondence estimation. The red curve, which represents the combination of key points and mutual strategy in Figure \ref{fig:global_performance}b, demonstrates that the correct matching ratio mostly falls within the 20\% to 50\% range, which is acceptable for robust initial pose estimation.

CryoAlign adopts two-stage alignment architecture to achieve precise pose estimation. The aforementioned key points based feature matching is utilized in the first stage, which provides a robust but relatively coarse pose. This stage serves as a foundation for the alignment process. In the second stage, CryoAlign shifts its focus to the initial sampling points after transformation, aiming to bring the two point clouds close enough. By combining these two stages, CryoAlign generates a more accurate superimposition of the density maps. Figure \ref{fig:global_performance}c demonstrates that the RMSD distribution of the two-stage alignment is more concentrated on the left three bars, indicating higher accuracy. Moreover, thanks to the initial pose estimation provided by the first stage, the second stage of point-based alignment requires less time to converge. Figure \ref{fig:global_performance}d presents the distribution of execution time for the alignment processes, revealing that the duration of the second stage is acceptable considering the improvement in accuracy.

\begin{figure}[H]
	\centering
	\includegraphics[width=0.85 \textwidth]{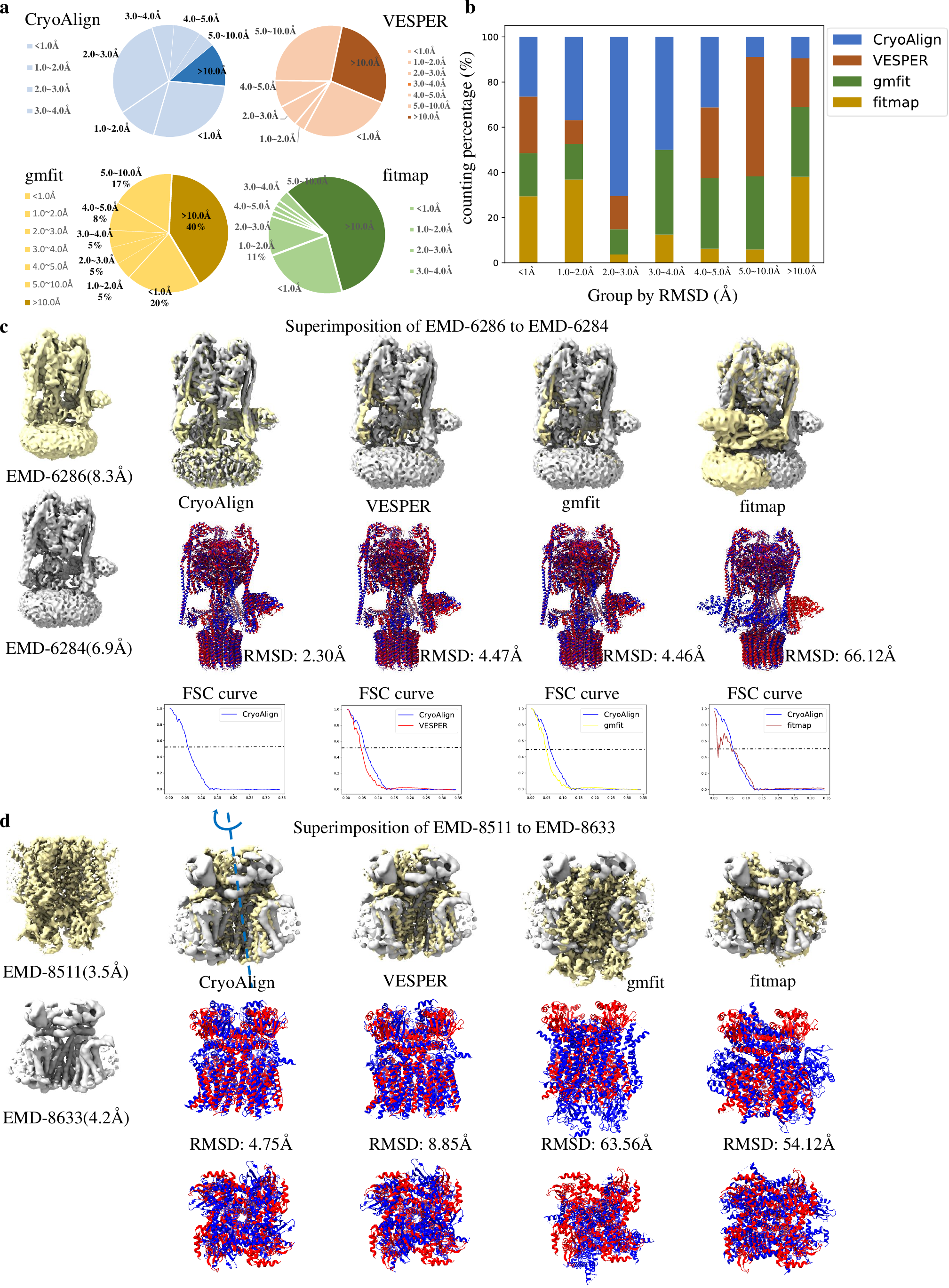}
	\caption{\textbf{Global alignment performance of compared methods.} \textbf{a} The RMSD distribution of the compared methods CryoAlign, VESPER, gmfit and fitmap. The sectors colored dark are the failure proportion (RMSD larger than 10\AA) of the methods, with CryoAlign/VESPER/gmfit/fitmap being 12\%/28\%/40\%/58\% respectively. Meanwhile, RMSD smaller than 3\text{\AA} can be considered as high quality alignment, with CryoAlign/VESPER/gmfit/fitmap being 69\%/36\%/30\%/35\% respectively. \textbf{b} The quantity statistical proportions for different RMSD groups. \textbf{c} The left example is the density map pair for the same state of Yeast V-ATPase (EMD-6286 and EMD-6284). There are little differences between the two maps. The alignment accuracy is evaluated by FSC curves on the right. The RMSD of CryoAlign/VESPER/gmfit/fitmap is 2.30/4.47/4.46/66.12\AA, respectively. \textbf{d} The right example is the density map pair for different states of Cyclic Nucleotide-Gated Ion Channel (EMD-8632 and EMD 8511). Accurate rotation estimation is in need here. The RMSD of CryoAlign/VESPER/gmfit/fitmap is 4.75/8.85/63.56/54.12\AA, respectively.}
	\label{fig:global_alignment}
\end{figure}

Figure \ref{fig:global_alignment} a, b collect RMSD distributions of CryoAlign and other comparative methods VESPER, gmfit and fitmap in global alignment datasets. For VESPER, the sampling and initial rotation intervals are set to 5\text{\AA} and $10^{\circ}$ respectively. Gmfit was run with 20 Gaussians and parameter -maxsize 64, which are the settings in the Omokage map web server. For fitmap, we take 20 random poses as the initial placements. The pie charts provide an overview of the alignment results for different methods, with the dark sections representing the failure proportion (RMSD larger than 10Å) and the shallow sections representing acceptable results. And the histogram shows the proportions of comparative methods in the different RMSD intervals. The results from fitmap exhibit a highly polarized distribution, with a majority of cases falling into the $>$10.0Å and $<$2.0Å ranges. This indicates a strong dependence on the quality of the initial poses provided. Gmfit shows a relatively average distribution, but it has the smallest section in the $<$1.0Å range, suggesting lower accuracy due to its blurred Gaussian representation. Compared to gmfit and fitmap, VESPER exhibits significant improvement in success rate, reducing the failure proportion to 28\%. However, its fixed exhaustive interval leads to RMSD values primarily falling within the 3.0$\sim$10.0\text{\AA} range. In contrast, CryoAlign achieves the lowest failure ratio and highest accuracy, with the majority of RMSD values concentrated below 3.0\AA.

Table \ref{tab:global_dataset} summarizes the average RMSD for different resolution ranges and execution time of CryoAlign and other compared methods. The first column of the table represents the resolution range of the two input density maps, and "Cross resolution" indicates that the maps have different resolution ranges. It is important to note that we define an RMSD larger than 10\text{\AA} as an alignment failure. Among the methods evaluated, fitmap has the lowest average RMSD in the first two rows, but it also has a failure rate of around 50\%. This is because fitmap heavily relies on the given initial poses, which are based on the domain knowledge of researchers. Without accurate initial poses, fitmap tends to produce poor alignment results. CryoAlign achieves the second-lowest average RMSD after fitmap and the lowest failure rate, demonstrating the superiority of its local spatial structural feature descriptors architecture. We also collected execution time information for the point generation process and the alignment stage of the four methods. Gmfit models the density maps by the combinations of multiple Gaussian kernels, which provide a rough representation of the 3D shape. Because gmfit utilizes a relatively minimal number of weights and parameters, it executes the fastest but with relatively lower accuracy. The execution time of fitmap mainly depends on the number of initial poses, and in our experiments, using 20 initial poses strikes a balance between accuracy and efficiency. Compared to other methods, CryoAlign takes considerable time in point extraction due to the additional key point descriptor computation. However, at the alignment stage, CryoAlign executes much faster. This is mainly because VESPER uses an exhaustive search method with repetitive operations, while CryoAlign directly estimates the transformation matrix. In summary, CryoAlign outperforms the compared methods in terms of both accuracy and efficiency in global alignment, with comprehensive consideration of both accuracy and efficiency.

\begin{table}[h]
	\caption{Alignment evaluation in global dataset}
	\label{tab:global_dataset}
	\centering
	\setlength{\tabcolsep}{4.mm}{
            \begin{threeparttable}
		\begin{tabular}{lcccc}
			\toprule
			Res. range              & CryoAlign(\AA)/failure & VESPER(\AA)/failure & gmfit(\AA)/failure & fitmap(\AA)/failure \\
			\textless{}5\AA   & \underline{1.69}/\textbf{18.4\%}   & 2.853/\underline{25.71\%}  & 3.01/37.14\%  &   \textbf{0.78}/48.57\%         \\
			5.0$\sim$10.0\AA      & \underline{2.88}/\textbf{6.25\%}   & 5.09/\underline{25\%}  & 7.59/25\% &   \textbf{0.82}/50\%      \\
			Cross res.      & \textbf{2.23}/\textbf{0\%}   & 4.47/\underline{30.77\%}  & \underline{3.58}/46.15\%  &  3.9/61.54\%           \\
			\midrule
			Time            & CryoAlign(s) & VESPER(s) & gmfit(s) & fitmap(s) \\
			extract points    & 18.9     & 3.1    &  5.35  &  -      \\
			alignment         & 0.94     & 202.5  &  0.213	& 60.12 \\
			total time		  & 19.84    & 205.6  &  5.56   & 60.12 \\
			\bottomrule
		\end{tabular}
            \begin{tablenotes}
                \footnotesize
                \item Three are two metrics calculated in the alignment evaluation, average RMSD value and failure ratio, whose the best ones are marked in bold and the second best ones are underlined. For RMSD, the smaller value means the better alignment accuracy; for failure ratio, the smaller values indicates the higher stability.
            \end{tablenotes}
            \end{threeparttable}
	}
\end{table}

\subsection{Examples of global alignment}
For a direct and fair comparison, we collected test examples of different resolutions in VESPER (Table 2 in the manuscript). Table \ref{tab:global_example} summarizes the RMSD of the best superimposition achieved by CryoAlign compared to VESPER, gmfit, and fitmap. It is important to note that the parameter combination used for VESPER is set to (1\text{\AA}, $10^{\circ}$), and the performances of gmfit and fitmap are directly taken from the recommendations from their paper. In cases where the inputted maps have the same resolution range (either $<5$\text{\AA} or $5\sim10$\text{\AA}), CryoAlign achieves results that are closest to the ground truth superimposition. Even when given maps have different resolutions, CryoAlign still provides acceptable pose estimation. This comparison demonstrates the effectiveness of CryoAlign in achieving accurate and reliable alignment results, especially when dealing with maps of the same resolution range. Meanwhile, it also showcases the robustness of CryoAlign in handling different resolutions and its ability to estimate accurate poses even in challenging scenarios.

\begin{table}[h]
	\caption{Examples for global map alignment}
	\label{tab:global_example}
	\centering
	\setlength{\tabcolsep}{2.mm}{
        \begin{tabular}{llrccccc}
            \toprule
            Res. range    & Map1 IDs       & Map2 IDs       & PDB RMSD(\AA) & \multicolumn{4}{l}{RMSD} \\ \cmidrule{5-8}
            &                &                &          & CryoAlign & VESPER(1\AA) & gmfit & fitmap \\ \cmidrule{5-8} 
            \textless{}5\AA & 3240/5fn5      & 2677/5a63      & 1.91     & \textbf{1.53}    & \underline{2.21}       & 2.63  & 2.90   \\
            & 8881/5wpq      & 8764/5w3s      & 2.08     & \textbf{0.68}    & \underline{1.12}       & 1.19  & 56.99  \\
            & 9515/5gjw      & 6475/3jbr      & 4.37     & \textbf{0.72}    & \underline{2.31}       & 2.95  & 97.48  \\
            5-10\AA         & 8744/5vy8      & 8267/5kne      & 3.44     & \textbf{0.39}    & \underline{0.86}       & 2.30  & 73.67  \\
            & 6284/3j9t      & 8724/5vox      & 5.13     & \underline{1.06}    & 2.79       & 5.05  & \textbf{1.04}   \\
            & 3342/5fwm      & 3341/5fwl      & 1.45     & \textbf{0.51}    & \underline{0.56}       & 3.60  & 4.98   \\
            Cross res.     & 8784/5w9i(3.6) & 8789/5w9n(5.0) & 8.33     & \textbf{0.05}    & \underline{2.84}       & 4.69  & 79.51  \\
            & 9515/5gjw(3.9) & 6476/3jbr(6.1) & 4.37     & \underline{4.31}    & \textbf{3.12}       & 6.06  & 64.19  \\
            & 3238/5fn3(4.1) & 2678/5a63(5.4) & 0.68     & \textbf{2.73}    & 3.34       & \underline{3.22}  & 3.68
            \\
            \bottomrule
        \end{tabular}
	}
\end{table}

Furthermore, Figure \ref{fig:global_alignment} c, d show two classical examples of global alignment. The first example involves a density map pair representing the same state of Yeast V-ATPase (EMD-6286 and EMD-6284). These maps are nearly identical, with only minor differences caused by molecular dynamics or imaging variations. In this case, the accuracy of translation parameter estimation plays a crucial role in alignment accuracy. Both CryoAlign and VESPER show excellent visual performance in terms of superimposition. However, the difference in RMSD is mainly reflected in the Fourier Shell Correlation (FSC) curve. The FSC figure on the right side of the example illustrates that the blue curve, representing CryoAlign, is consistently positioned to the right of the red curve, indicating more accurate alignment parameters. 
The second example involves a density map pair representing different states of the Cyclic Nucleotide-Gated Ion Channel (EMD-8632 and EMD-8511). These maps exhibit structural similarities but have significant contour differences. Additionally, there is a rotational invariance around an axis, which imposes higher requirements on rotation parameter estimation. For comparison, we provide two different viewing directions of the PDB atom model superimposition. The left view represents the ordinary viewing direction, while the right view represents the rotation axis view. From the ordinary viewing direction, both CryoAlign and VESPER demonstrate accurate translation parameter estimation. However, from the rotation axis viewing direction, we observe that VESPER exhibits a larger RMSD, indicating poor rotation parameter estimation. One possible reason for this discrepancy is that the approximate rotational invariance leads to mismatches in vectors, which only reflect the density trend of a single point. In contrast, CryoAlign utilizes the orientation distribution of local regions as features, allowing for more accurate estimation of rotation parameters.

\subsection{Local alignment accuracy}
\begin{figure}[H]
	\centering
	\includegraphics[width=0.8 \textwidth]{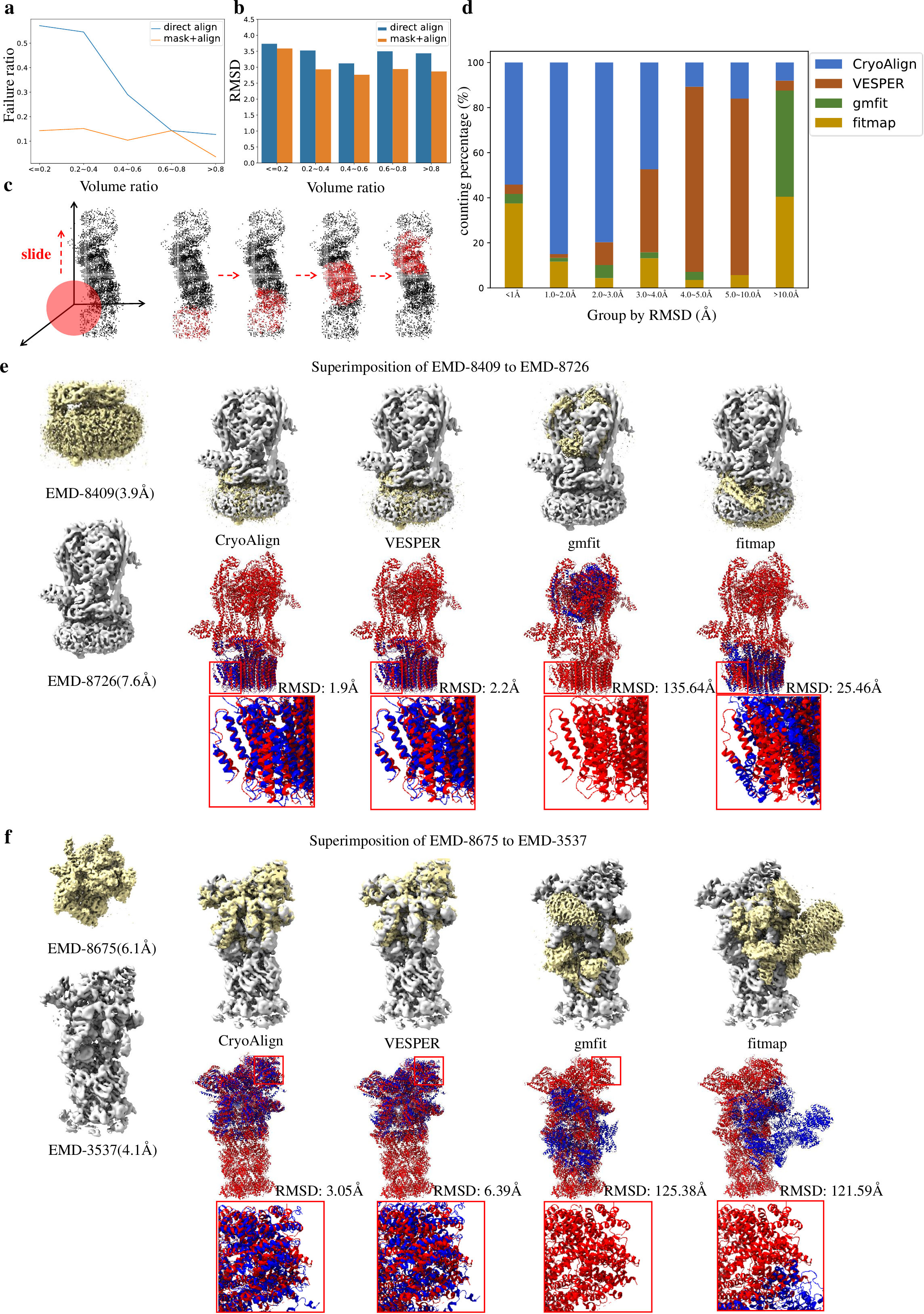}
	\caption{\textbf{Local alignment. a} The relation between failure probability and volume proportion of the smaller map to the larger one. The blue curve is direct alignment without cutting. The orange curve is multiple alignment with translational mask. \textbf{b} The alignment accuracy of two alignment strategy, direct alignment and multiple alignment with mask. \textbf{c} The sketch of translational mask. The mask moves in given interval along the axis and part of the larger point cloud is taken for alignment. The extracted points are labelled by red and the remaining ones are black. \textbf{d} The quantity statistical proportions for different RMSD groups. \textbf{e} The first example is superimposing the Vo region of the V-ATPase (EMD-8409) on the complete V-ATPase (EMD-8726). Although the volume ratio is smaller than 50\%, the distinct fence-like 3D structure makes EMD-8409 distinctive in EMD-8726. The RMSD of CryoAlign/VESPER/gmfit/fitmap is 1.9/2.2/135.64/25.46\AA, respectively. \textbf{e} The second example is to align 26S proteasome regulatory particle (EMD-8675) and 26S proteasome of Saccharomyces cerevisiae in presence of BeFx (EMD-3537). The volume ratio is about 50\%, but it is still difficult to align them using traditional methods. The RMSD of CryoAlign/VESPER/gmfit/fitmap is 3.05/6.39/125.38/121.59\AA, respectively.}
	\label{fig:local_combined}
\end{figure}

In the context of local alignment, it is important to consider the size difference between the inputted density maps. If the size of the smaller map occupies more than 40\% of the size of the larger map (volume ratio), the accuracy of feature matching remains similar to that of global alignment in most cases. However, if the size difference is too large, it becomes challenging for feature-based alignment to find an acceptable superimposition in a single attempt. Figure \ref{fig:local_combined}a illustrates the higher failure probability as the volume ratio decreases. This is because the candidate feature descriptors from the larger map can easily interfere with the smaller number of feature queries. To address accurate local alignment, CryoAlign treats it as a global retrieval problem within a small "dataset". It adopts a translational mask as a simple segmentation scheme for the larger point cloud, as shown in Figure \ref{fig:local_combined}c. The two-stage alignment process is then used to calculate a series of pose parameters. Based on this collection of parameters, CryoAlign measures the similarity scores across all superimpositions and selects the top one as the output. Moreover, Figure \ref{fig:local_combined}b demonstrates that this mask strategy not only helps in cases with low volume ratios to find the best superimposition but also improves alignment accuracy in cases with high volume ratios.

The average RMSD and failure information for local alignment are presented in Table \ref{tab:local_dataset}. In comparison to global alignment, both gmfit and fitmap exhibit higher failure ratios, ranging from 80\% to even 100\%. This highlights the difficulty of directly aligning two complete density maps in local alignment. However, fitmap achieves the best performance when provided with sufficient initial poses. On the other hand, VESPER utilizes an exhaustive search architecture to find the optimal parameters, making it still applicable in local alignment cases. Similarly, CryoAlign generates a series of candidate parameters based on the translational mask and selects the best one. This simple segmentation strategy effectively transforms the local alignment problem into multiple global alignment problems, thereby ensuring the accuracy of the feature matching stage to a certain extent. Furthermore, similar to global alignment, CryoAlign demonstrates lower average RMSD values, indicating superior performance compared to VESPER within the same sampling interval.

Two examples of local alignment are shown in Figure \ref{fig:local_combined} e, f. In the first example, we aim to superimpose the Vo region of the V-ATPase (EMD-8409) onto the complete V-ATPase (EMD-8726). Despite EMD-8409 occupying less than 40\% of the volume of EMD-8726, its distinct fence-like 3D structure makes it stand out within the complete V-ATPase map. Both CryoAlign and VESPER achieve high alignment accuracy, with RMSD values of approximately 2Å, significantly lower than the sampling interval of 5Å. Gmfit fails to accurately capture the local structures due to its rough Gaussian representations. Fitmap, despite accepting an approximate initial pose, also fails due to excessive focus on the overlapping region. Upon observing the enlarged PDB models, we can see that fitmap attempts to align the right side better while neglecting the left side. The second example involves the alignment of the 26S proteasome regulatory particle (EMD-8675) and the 26S proteasome of Saccharomyces cerevisiae in the presence of BeFx (EMD-3537). The failures of gmfit and fitmap demonstrate that when the smaller map occupies approximately 50\% of the larger one, it becomes challenging to align them using conventional methods. VESPER adopts an exhaustive search architecture to find an acceptable superimposition, but the fixed translation and rotation intervals limit its precision. In contrast, CryoAlign employs a correspondence-based method to estimate "sub-voxel" transformation parameters, resulting in a lower RMSD.

\begin{table}[h]
	\caption{Alignment evaluation on the local dataset}
	\label{tab:local_dataset}
	\centering
	\setlength{\tabcolsep}{4.mm}{
            \begin{threeparttable}
		\begin{tabular}{lcccc}
			\toprule
			Res. range              & CryoAlign(\AA)/failure & VESPER(\AA)/failure & gmfit(\AA)/failure & fitmap(\AA)/failure \\
			\textless{}5\AA   & \underline{3.55}/\underline{8.2\%}   & 6.07/\textbf{0.0\%}  & 8.05/92.6\%  &   \textbf{2.34}/91.8\%         \\
			5.0$\sim$10.0\AA      & \textbf{4.22}/\textbf{0.0\%}   & \underline{6.48}/\underline{7.1\%}  & 12.62/57.1\% &   6.55/85.7\%      \\
			Cross res.      & \underline{4.50}/\textbf{13.8\%}   & 7.02/\textbf{13.8\%}  & -/100\%  &  \textbf{4.15}/75.4\% \\
			\bottomrule
		\end{tabular}
            \begin{tablenotes}
                \footnotesize
                \item Three are two metrics calculated in the alignment evaluation, average RMSD value and failure ratio, whose the best ones are marked in bold and the second best ones are underlined. For RMSD, the smaller value means the better alignment accuracy; for failure ratio, the smaller values indicates the higher stability.
            \end{tablenotes}
            \end{threeparttable}
	}
\end{table}

\subsection{Application in map comparison}
Accurate alignment of density maps is an essential step in heterogeneity analysis or 3D classification. Existing softwares often employ cross-correlation-based methods to directly quantify voxel differences between maps. It usually works well when the maps are roughly pre-aligned or the differences are not significant enough. In fact, cross-correlation methods still encounter issues arising from inadequate initial poses. As a point cloud based approach, CryoAlign might not provide the same level of precise superimposition as cross-correlation methods, due to information loss resulting from the point sampling process. But CryoAlign has the ability to achieve a sufficiently close map superimposition, which could potentially serve as an initial pose for subsequent refinement processes. 

\begin{figure}[H]
	\centering
	\includegraphics[width=0.9 \textwidth]{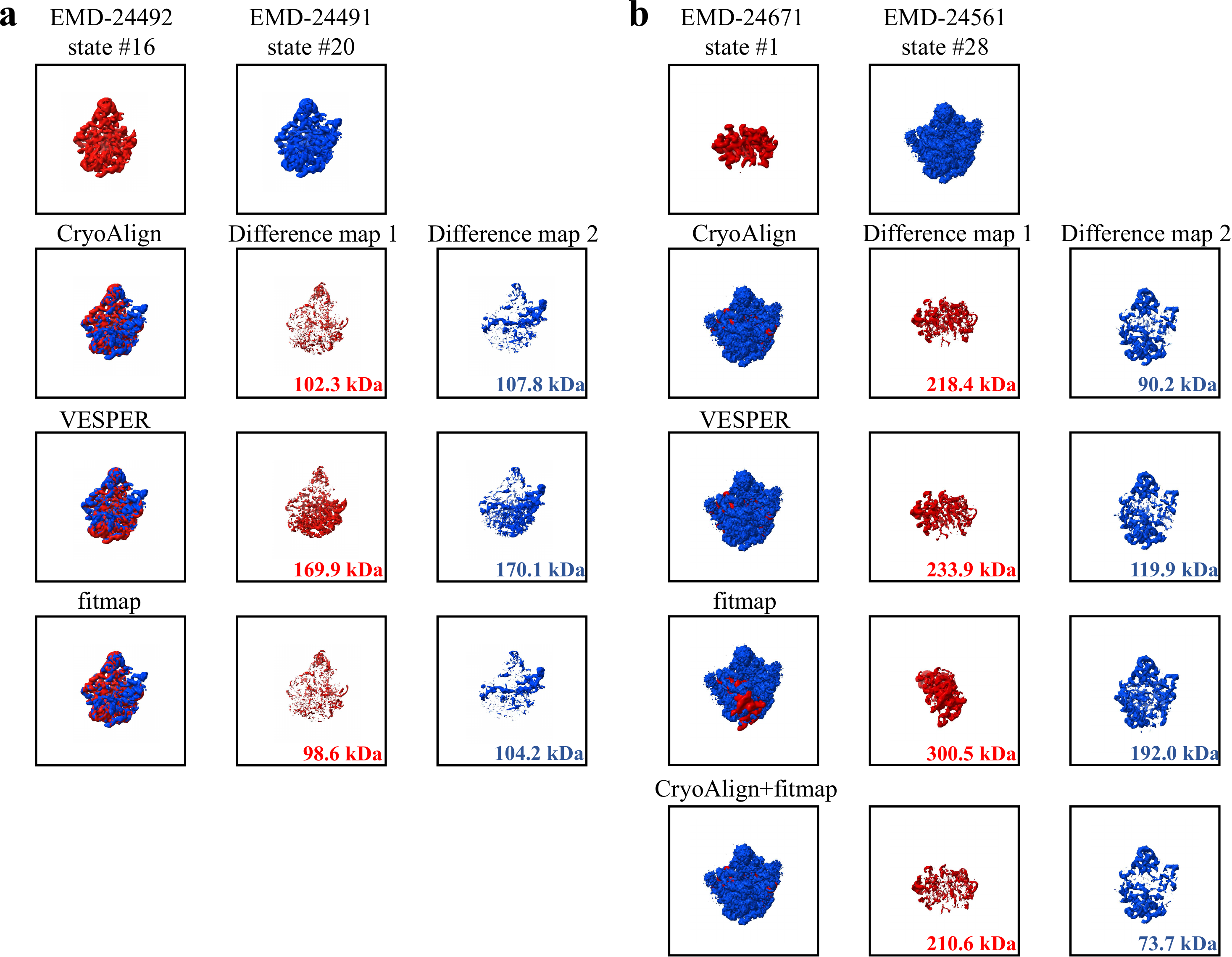}
	\caption{\textbf{Examples for heterogeneity analysis.}}
	\label{fig:heterogeneity1}
\end{figure}

In Figure \ref{fig:heterogeneity1}, we present two examples showing different states of bL17-limited ribosome assembly intermediates \cite[]{rabuck2022quantitative}. Figure \ref{fig:heterogeneity1} a illustrates a comparison between state \#16 (EMD-24492) and state \#20 (EMD-24491). These two states are quite similar, with the primary distinction being an area in the upper right corner. We compute the difference maps for both scenarios: source map - target map and target map - source map. The differences are defined as changes in molecular weight, which directly correspond to the voxel-based difference densities and are calculated using 0.81 Da/$\AA^3$. Notably, CryoAlign achieves a comparable superimposition to fitmap, while VESPER produces a less accurate result. In Figure \ref{fig:heterogeneity1}b, we analyze the comparison between state \#1 (EMD-24671) and state \#28 (EMD-24561). Substantial differences exist between the two maps, posing a challenge for cross-correlation based methods like fitmap. Compared to VESPER, CroAlign offers a better superimposition, which can serve as a acceptable initial position for the subsequent refinement. In the "Difference map 2" column, the molecular weight of CryoAlign is significantly lower than the weight of VESPER. Furthermore, the combination of CryoAlign and fitmap yields the lowest weight, demonstrating the feasibility of integrating these two methods.

\begin{figure}[H]
	\centering
	\includegraphics[width=0.9 \textwidth]{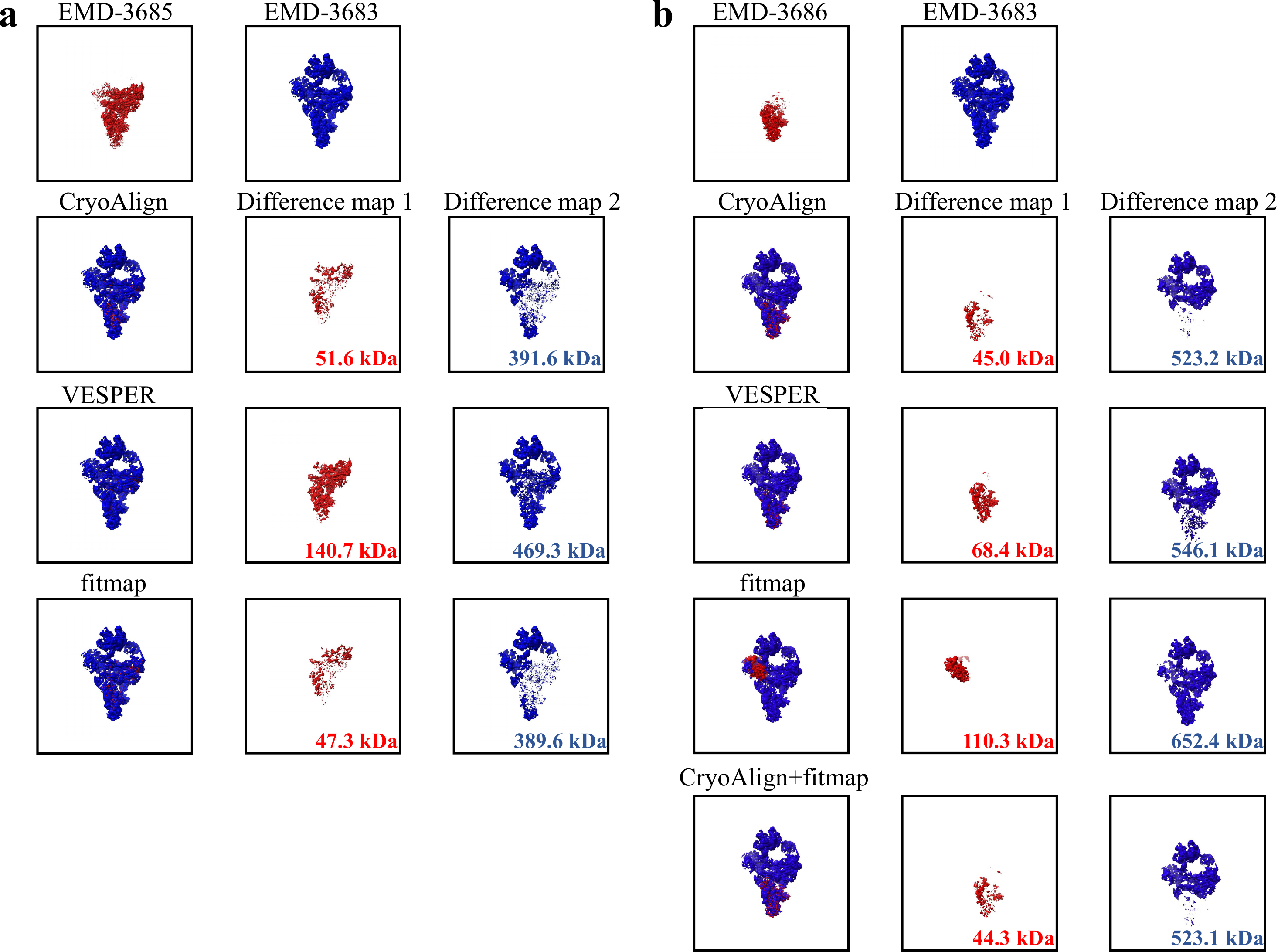}
	\caption{\textbf{Examples for heterogeneity analysis.}}
	\label{fig:heterogeneity2}
\end{figure}

We provide an additional example of pre-catalytic spliceosome for comparison in Figure \ref{fig:heterogeneity2}. VESPER typically estimates a roughly accurate transformation parameter, limited by its pre-given rotational and translational intervals. Fitmap outperforms where most regions of the two maps exhibit the same. However, it also fails when one map occupies a significantly smaller portion within the other map. In contrast, CryoAlign achieves alignment accuracy comparable to fitmap, as demonstrated in Figure \ref{fig:heterogeneity2}a. Moreover, CryoAlign proves effective in providing a robust initial position even in challenging scenarios, as depicted in Figure \ref{fig:heterogeneity2} b.

\subsection{Application in atomic model fitting}
Local alignment plays a crucial role in the assembly of single chains in protein complex atom modeling. To facilitate this process, we gather a set of density maps representing protein complexes along with their associated PDB entries. From each fitted PDB atom model, we extract all single chains present. For every single chain, we simulate a corresponding density map using the "molmap" command in Chimera, ensuring that the resolution matches that of the target complete map. To achieve higher alignment accuracy, we set the initial sampling interval to 3Å for both CryoAlign and VESPER. This choice is motivated by the small size of the single protein chains, where a smaller sampling interval can provide more detailed structural information.

\begin{figure}[H]
	\centering
	\includegraphics[width=0.7 \textwidth]{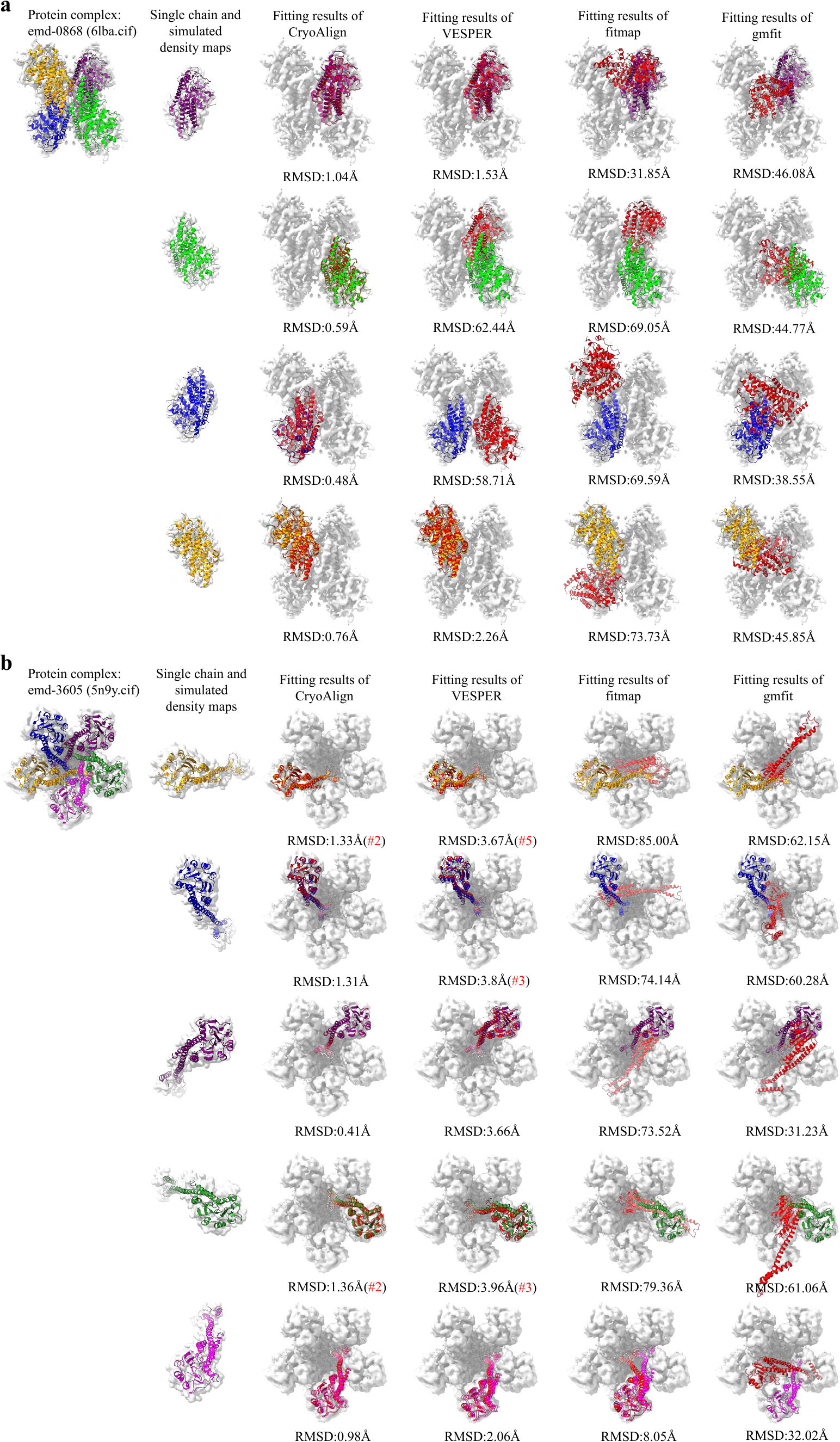}
	\caption{\textbf{Examples for atomic model fitting.} \textbf{a} Chain structure fitting of pentameric ZntB transporter (EMD-3605, PDB: 5n9y), which consists of five single chains A, B, C, D and E. For rotational invariance caused by similarity of chains, we collect the top-scoring five of CryoAlign or VESPER as candidates and select the best one. The red "\#2" besides the RMSD value is the ranking of the best superimposition in the candidate list. \textbf{b} Chain structure fitting of kinase domain-like (MLKL) protein (EMD-0868, PDB: 6lba), which consists of four single chains A, B, C and D. Noted that, if RMSD value is large but no ranking is listed aside for CryoAlign or VESPER, it indicates that none of the top-scoring five parameters resulted in a successful alignment.}
	\label{fig:huang_example}
\end{figure}

We present two representative examples of atomic model fitting using CryoAlign. The first example involves the pentameric ZntB transporter (EMD-3605, PDB id: 5n9y), which consists of five single chains labeled A to E (Figure \ref{fig:huang_example} a). Due to the structural similarity among the five chains, they exhibit a certain degree of rotation invariance. In order to account for this invariance, we provide the top five scoring parameters and indicate the rank of the best superimposition. In Figure \ref{fig:huang_example} a, the rank is denoted by "(\#2)" next to the RMSD value in red. If no ranking information is given, the RMSD is calculated based on the top-scoring alignment (i.e. in default, the RMSD is calculated in the ranking first alignment). In this example, gmfit and fitmap generally fail to produce satisfactory results. While VESPER finds acceptable alignment parameters, the rankings of the three chain results are low. This is primarily due to the fixed intervals used in the exhaustive search method, which results in a lack of discrimination among the top candidate alignments. CryoAlign, on the other hand, utilizes feature-based alignment to estimate more consistent and robust initial poses, leading to higher rankings for the chain results compared to VESPER. The second example (Figure \ref{fig:huang_example} b) involves the kinase domain-like (MLKL) protein (EMD-0868, PDB id: 6lba). It should be noted that if no ranking is provided alongside the RMSD, it means that none of the top five scoring parameters yielded successful alignments. For instance, the second and third rows of VESPER in the example demonstrate its inability to find the correct position due to the rotational invariance. Furthermore, using the same sampling interval, CryoAlign achieves more accurate alignment performance in terms of RMSD compared to VESPER.

We have collected the rest atomic model fitting results in Table \ref{tab:local_atom}, which provides alignment accuracy for each single chain. The RMSD values ranging from 30 to 60\text{\AA} clearly demonstrate the inadequacy of gmfit and fitmap without any manual intervention. VESPER, on the other hand, is capable of identifying the correct superimposition from the top-scoring ten candidates. However, due to the lack of distinction in transformation parameters, the correct alignment often receives low rankings. In the case of chain E in EMD-0440 and chain D in EMD-4400, the acceptable alignment is not even present in the top ten list. In comparison, CryoAlign achieves higher alignment accuracy as indicated by lower RMSD values. Moreover, the correct parameters tend to rank higher in CryoAlign's candidate lists, further highlighting the superiority of the feature-based alignment approach.

\begin{table}[h]
	\caption{Alignment evaluation in atomic model fitting}
	\label{tab:local_atom}
	\centering
	\setlength{\tabcolsep}{4.mm}{
		\renewcommand\arraystretch{0.8}
		\begin{tabular}{l|c|cccc}
			\toprule
			Protein complex                              & chain & CryoAlign/\AA & VESPER/\AA    & gmfit/\AA & fitmap/\AA \\
			\midrule
			\multirow{2}{*}{emd\_0287(6hv8.cif)} & A     & \textbf{0.94}    & \underline{3.56}      & 38.46 & 54.11  \\
												 & B     & \textbf{0.67}    & \underline{3.68}      & 50.83 & 46.82  \\
			\midrule
			\multirow{2}{*}{emd\_0346(6h52.cif)} & A     & \textbf{0.47}    & \underline{2.46}      & 75.59 & 0.96   \\
												 & B     & \textbf{0.50}    & \underline{2.58}      & 94.86 & 1.27   \\
			\midrule
			\multirow{3}{*}{emd\_0440(6nd1.cif)} & A     & \textbf{0.41}    & \underline{2.72}      & 36.65 & 67.09  \\
												 & B     & \textbf{0.76}    & \underline{3.42}      & 33.21 & 50.99  \\
												 & E     & \textbf{0.82}    & \underline{43.93}     & 55.35 & 76.62  \\
			\midrule
			\multirow{4}{*}{emd\_0868(6lba.cif)} & A     & \textbf{1.04}    & \underline{1.54}      & 46.08 & 31.85  \\
												 & B     & \textbf{0.59}    & \underline{3.70}(\#9) & 44.77 & 69.05  \\
												 & C     & \textbf{0.48}    & 58.71     & \underline{38.55} & 69.59  \\
												 & D     & \textbf{0.76}    & \underline{2.62}      & 45.85 & 73.73  \\
			\midrule
			\multirow{3}{*}{emd\_3340(5fwp.cif)} & A     & \textbf{1.35}      & \underline{1.92}      & 63.25 & 63.92  \\
												 & B     & \textbf{2.01}(\#5) & \underline{4.11}(\#5) & 21.55 & 41.89  \\
												 & K     & \textbf{4.47}      & \underline{5.00}      & 37.19 & 26.71  \\
			\midrule
			\multirow{5}{*}{emd\_3605(5n9y.cif)} & A     & \textbf{1.33}(\#2) & \underline{3.67}(\#5) & 62.15 & 85.00  \\
												 & B     & \textbf{1.31}      & \underline{3.80}(\#3) & 60.28 & 74.14  \\
												 & C     & \textbf{0.41}      & \underline{3.66}      & 31.23 & 73.52  \\
												 & D     & \textbf{1.36}(\#2) & \underline{3.96}(\#3) & 61.06 & 79.36  \\
												 & E     & \textbf{0.98}      & \underline{2.06}      & 32.02 & 8.05   \\
			\midrule
			\multirow{2}{*}{emd\_3861(5oyg.cif)} & A     & \textbf{0.48}      & \underline{4.39}      & 53.62 & 61.33  \\
												 & B     & \textbf{0.27}      & \underline{3.86}      & 46.85 & 55.69  \\
			\midrule
			\multirow{5}{*}{emd\_4400(6i2t.cif)} & A     & \textbf{1.27}(\#3) & \underline{2.32}      & 63.52 & 75.85  \\
												 & B     & \textbf{1.03}      & \underline{2.30}(\#2) & 63.57 & 42.64  \\
												 & C     & \textbf{1.03}      & \underline{3.96}(\#2) & 35.02 & 80.75  \\
												 & D     & \textbf{0.95}      & 73.82     & \underline{59.62} & 112.40 \\
												 & J     & \textbf{4.39}      & \underline{9.29}      & 65.23 & 53.25 
			\end{tabular}
	}
\end{table}

\section{Discussion}
In this study, we introduced CryoAlign, a highly accurate method for aligning electron microscopy (EM) density maps at both global and local levels. CryoAlign operates by transforming the inputted maps into 3D points and leveraging local spatial structual feature descriptors to effectively capture the underlying structural information. The alignment process in CryoAlign is conducted in two stages. In the first stage, CryoAlign employs clustering-based key point extraction and mutual feature matching techniques to establish correspondences between the extracted key points. This enables CryoAlign set a solid foundation for achieving fast and robust superimposition. In the second stage, CryoAlign focuses on establishing correct point-to-point correspondences between the sampled points. By carefully building these correspondences, CryoAlign calculates the final transformation parameters, resulting in a highly precise superimposition.

CryoAlign surpasses existing methods in terms of alignment accuracy for global alignment tasks, while maintaining a good execution time. By achieving more precise superimposition of density maps, CryoAlign enables researchers to identify and analyze differences or changes between two maps, leading to a better understanding of biological structures. While the parameter settings used in the experiment results demonstrate the superior alignment performance of CryoAlign, it is worth noting that these settings are not necessarily optimized for all tasks or imaging environments. Users have the flexibility to explore different parameter configurations based on their specific requirements. In addition to alignment accuracy, CryoAlign offers a scoring function that measures the similarity between two maps. This scoring function can be used in map retrieval tasks, allowing researchers to search for maps with similar characteristics or features.

For local alignment, CryoAlign employs the local spatial structural feature descriptors based alignment combined with a segmentation approach. The simple segmentation strategy using translational masks has demonstrated its effectiveness in experiments, but it may suffer from redundancy. By incorporating domain knowledge and developing a more advanced segmentation scheme, CryoAlign has the potential to achieve even faster and more accurate results in local alignment tasks. Local map alignment plays a crucial role in the subunit assembly of protein macromolecular atom modeling. Since identical single chains may exist in the structure, CryoAlign provides multiple transformation candidates ranked by similarity scores. Users can evaluate each alternative superimposition and select the most suitable one based on their domain knowledge and expertise.

In conclusion, CryoAlign offers a robust and accurate alignment solution for EM density maps with a resolution higher than 10\text{\AA}. Its capabilities in both global and local alignment make it a valuable tool for studying and analyzing structural biology EM maps. CryoAlign's ability to accurately superimpose maps enables researchers to gain deeper insights into the structural details and variations present in the maps.

\section{Methods} 
\subsection{Point cloud generation} CryoAlign starts by converting the input density map into a point cloud through uniform sampling, assigning density vectors using the mean shift equation. It then identifies key points within the point cloud using clustering techniques and computes local spatial structural feature descriptors. These key points and feature descriptors are utilized in the subsequent alignment stages to achieve accurate alignment.

\textbf{Initial density-based points generation.} The successful application of VESPER demonstrates that the intensive unit vectors have the ability to capture the local structures of density maps. CryoAlign regards the uniformly sampled grid points as the point cloud and calculates unit vectors as the "density vectors" for these points. At each grid point $x_i(i=1,...,N)$ with a density value that no less than author-recommended contour level, the unit vector located at $x_i$ is $\vec{\frac{y_i-x_i}{|y_i-x_i|}}$, where $y_i$ is defined as follows:
\begin{align}
    & y_i = \frac{\sum_{n=1}^N k(x_i-x_n)\Phi(x_n)x_n}{\sum_{n'=1}^N k(x_i-x_n')\Phi(x_n')} \label{eq:meanshift}\\
    & k(p) = \exp(-1.5|\frac{p}{\sigma}|^2)
\end{align}
where $\sigma$ is a bandwidth and $\Phi(x_n)$ is the density value of the grid point $x_n$.

\textbf{Clustering-based key points and descriptors extraction.} 
In EM maps, the density value corresponds to the integration of density functions related to atoms, and regions with high density can be indicative of protein backbones. CryoAlign employs the mean shift algorithm, a non-parametric clustering method, to effectively identify these dense regions in the map. By iteratively applying Equation \ref{eq:meanshift} until convergence, CryoAlign determines the local density maximum points by computing the cluster centers.
\begin{align}
    y_i^{t+1} = \frac{\sum_{n=1}^N k(y_i^t-x_n)\Phi(x_n)x_n}{\sum_{n'=1}^N k(y_i^t-x_n')\Phi(x_n')}
\end{align}
To enhance the representation capability and reduce the size of the point cloud, CryoAlign incorporates the DBSCAN (Density-Based Spatial Clustering of Applications with Noise) algorithm\cite[]{ester1996density}. This algorithm clusters points that are located within a specified threshold distance, typically equivalent to the sampling space. By applying DBSCAN, CryoAlign groups nearby points together, effectively reducing the redundancy and capturing the essential structural information in a more compact form. The remaining points serve as key points for subsequent alignment stages.

Based on identified key-points and initial points assigned with "density vectors", CryoAlign proceeds to calculate density based Signature of Histograms of Orientations (SHOT) feature descriptors \cite[]{salti2014shot} for each key point (see Supplementary). To calculate the modified SHOT descriptors, CryoAlign examines the local neighborhood points surrounding each key point. The orientations of the assigned density vectors at these neighboring points are quantized into discrete bins, and a histogram is constructed to collect the distribution of these orientations. This histogram effectively summarizes the local geometric characteristics of the density map in a concise and informative manner.

\subsection{Two-stage alignment}
After the sampling and clustering stages, two inputted density maps are efficiently transformed into source (moving) point cloud and key points, denoted as $\{S_i, S_i^{key}\}$, and target (fixed) point cloud and key points, denoted as $\{T_j, T_j^{key}\}$.
In the first stage of alignment, CryoAlign utilizes a feature-based approach to estimate the initial transformation parameters. This involves collecting the key points and their corresponding feature descriptors from both the source and target point clouds. To efficiently reduce the size of the candidate set, CryoAlign employs a bidirectional nearest point matching strategy. This strategy assigns a binary value, denoted as $\delta(i, j)$, to each pair of key points, indicating whether they should be considered as a potential match. When $\delta(i, j) = 1$, it means that the corresponding feature pair between key point $S_i$ and key point $T_j$ is considered as a valid match. On the other hand, when $\delta(i, j) = 0$, it means that the corresponding feature pair is discarded.
\begin{align}
	\delta(i,j)=NN(S_i^{key},T_j^{key})\  \& \ NN(T_j^{key},S_i^{key})
\end{align}
where NN($\cdot$) determines whether the latter point is the nearest one to the former point in the feature domain. In other words, CryoAlign compares the Euclidean distances between the feature descriptors of key point $S_i^{key}$ and all the feature vectors of key points in the target point cloud. If the feature descriptors of key point $S_i$ has the smallest distance to a certain key point in the target point cloud, then that key point is considered as the nearest neighbor (NN) of $S_i^{key}$.
Given the filtered feature point correspondences $\{S_i^{key}, T_i^{key}\}_{i=1}^M$, truncated least squares estimation and semidefinite relaxation (TEASER)\cite[]{Yang20troteaser} are used to estimate the initial rigid transformation parameters.
\begin{align}
	\min_{R\in SO(3),t\in \mathbb{R}}\sum_{i=1}^M\min(||T_i^{key}-RS_i^{key}-t||^2, \epsilon^2)
\end{align}

The feature-based method provides a rough initial superimposition, while the point-based method aims to align the point clouds more closely. Taking into account the different distributions of the point clouds, CryoAlign utilizes the sparse-icp algorithm\cite[]{bouaziz2013sparse} in the second stage. This algorithm replaces the L2 norm with the Lp norm (where p<1), allowing for a higher tolerance for outliers. Unlike the first stage that focuses on key point pairs, in the second stage, CryoAlign considers the initial point pairs $\{S_i, T_i\}_{i=1}^N$ generated by the nearest neighbor algorithm in 3D space. 
\begin{align}
	\min_{R\in SO(3),t\in \mathbb{R}} \sum_{i}^N \Vert T_i-RS_i-t\Vert_2^p + I_{SO(3)}(R)
\end{align}
where p $<$ 1 and $I_{SO(3)}$ constraints for the rotation matrix R.

\subsection{Similarity measuring function}
The similarity measuring function in CryoAlign is based on the aligned point clouds. Once the point clouds are transformed using the estimated alignment parameters, they are effectively superimposed. The similarity between the transformed point clouds $\{S_i\}$, $\{T_j\}$, along with their corresponding density vectors $\{u_i\}$, $\{v_j\}$, is measured as follows:
\begin{align}
	Similairty(S, T) &= (1 - D_{JS}(S|T)) * \frac{\sum_k^N I(u_k, v_k)}{N} \\
	I(u_k, v_k) &= 
	\left\{
		\begin{aligned}
			\quad1&, \quad\quad\quad\; &u_k * v_k > \epsilon \\
			\quad0&, &otherwise
		\end{aligned}
	\right.
\end{align}
where $D_{JS}(\cdot)$ is the Jensen-Shannon Divergence, measuring global similarity of the spatial distributions; $N$ in the denominator represents the number of overlapped point pairs; $I(\cdot, \cdot)$ is an indicator function, evaluating whether the dot product of two vectors is greater than a predefined threshold $\epsilon$.

\subsection{Exploration of local spatial features}
The combinations of keypoint detectors and feature descriptors are indeed crucial for achieving fast and effective initial alignment. There are several popular combinations available, such as keypoint detectors: 3D Harris \cite[]{sipiran2011harris}, 3D SIFT (Scale-invariant feature transform \cite[]{lowe2004sift,rusu2011pcl}), ISS (Intrinsic Shape Signatures \cite[]{zhong2009iss}); feature descriptors: SHOT (Signatures of Histograms of OrienTations \cite[]{salti2014shot}), FPFH (Fast Point Feature Histograms \cite[]{rusu2009fpfh}), PFH (Point Feature Histograms \cite[]{rusu2008pfh}), 3DSC (3D Shape Context \cite[]{kortgen20033dsc}), USC (Unique Shape Context \cite[]{tombari2010usc}), ROPS (Rotational Projection Statistics \cite[]{guo2013rops}). These algorithms are all computed with PCL library \cite[]{rusu2011pcl}. In the case of CryoAlign, density vectors are utilized as the geometry attribute for each point, replacing the commonly used surface normals in point cloud processing. Comparing density vectors with surface normals is also an important aspect.

In the supplementary material, we provide a comprehensive analysis of the aforementioned combinations compared to CryoAlign's approach using the global alignment dataset. The analysis includes evaluations of surface normals and density vectors for their orientation consistency in corresponding points, as measured by cosine distances. Meanwhile, the performance of correspondence establishment is assessed for different combinations of keypoint detectors and descriptors through metrics such as the proportion of correct feature matching and average precision. Different feature matching strategies, including direct nearest neighbor and mutual feature matching, are also tested for all combinations.
Through our analysis, we affirm that density vectors provide a better representation of geometry attributes compared to surface normals. The clustering-based keypoint extraction method also demonstrates superior performance. These methods take into consideration the physical meaning of density values, making them more applicable for density maps. To encode geometry attributes into a feature vector for each keypoint, we utilize the SHOT descriptor architecture, resulting in a 352-dimensional feature representation of local structures. The experiments detailed in the supplementary material demonstrate that the SHOT architecture exhibits robustness and accuracy, particularly when used in the mutual feature matching strategy.

\section{Availability of data and materials}
The datasets of EM maps for global or local alignment are provided in Supplementary data. The EM maps and fitted PDB entries can be downloaded from EMDB and PDB, respectively.

\section{Availability of code}
The CryoAlign program is freely available for academic use via *****.

\section{Funding}
This research was supported by the National Key Research and Development Program of China [2021YFF0704300], the King Abdullah University of Science and Technology (KAUST) Office of Research Administration (ORA) under Award No URF/1/4352-01-01, REI/1/5234-01-01, and REI/1/5414-01-01, the National Natural Science Foundation of China Projects Grant [62072280, 61932018, 62072441, T2225007 and 32241027], the Natural Science Foundation of Shandong Province ZR2023YQ057, and the Natural Science Foundation of Ningxia Province 2023AAC05036.

\bibliographystyle{natbib}
\bibliography{ref}

\end{document}